\documentclass[runningheads,a4paper]{llncs}

\usepackage{amsmath}
\usepackage{color}
\usepackage{graphicx}
\usepackage{array}
\usepackage{subfigure}
\usepackage{url}

\newcolumntype{C}[1]{>{\centering\let\newline\\\arraybackslash\hspace{0pt}}m{#1}}

\begin{document}

\mainmatter  

\title{Multi-Context Deep Network for Angle-Closure Glaucoma Screening in Anterior Segment OCT}

\author{Huazhu Fu$^1$ 
	\and Yanwu Xu$^2$\footnote{corresponding author} 
	\and Stephen Lin$^3$ 
	\and Damon Wing Kee Wong$^1$ 
	\and Baskaran Mani$^4$
	\and Meenakshi Mahesh$^4$
	\and Tin Aung$^4$
	\and Jiang Liu$^5$}

\authorrunning{H. Fu et al.}
\institute{$^1$ Institute for Infocomm Research, A*STAR, Singapore\\
	$^2$ Guangzhou Shiyuan Electronic Technology Company Limited, China\\
	$^3$ Microsoft Research, Beijing, China\\
	$^4$ Singapore Eye Research Institute, Singapore\\
	$^5$ Cixi Institute of Biomedical Engineering, CAS, China\\
	\url{https://hzfu.github.io/proj_glaucoma_asoct.html}.
}

\toctitle{Lecture Notes in Computer Science}
\tocauthor{Authors' Instructions}
\maketitle

\begin{abstract}

A major cause of irreversible visual impairment is angle-closure glaucoma, which can be screened through imagery from Anterior Segment Optical Coherence Tomography (AS-OCT). Previous computational diagnostic techniques address this screening problem by extracting specific clinical measurements or handcrafted visual features from the images for classification. In this paper, we instead propose to learn from training data a discriminative representation that may capture subtle visual cues not modeled by predefined features. Based on clinical priors, we formulate this learning with a presented Multi-Context Deep Network (MCDN) architecture, in which parallel Convolutional Neural Networks are applied to particular image regions and at corresponding scales known to be informative for clinically diagnosing angle-closure glaucoma. The output feature maps of the parallel streams are merged into a classification layer to produce the deep screening result. Moreover, we incorporate estimated clinical parameters to further enhance performance. On a clinical AS-OCT dataset, our system is validated through comparisons to previous screening methods.

\end{abstract}

\section{Introduction}

\begin{figure}[!t]
	\begin{center}
		\includegraphics[width=0.95\linewidth]{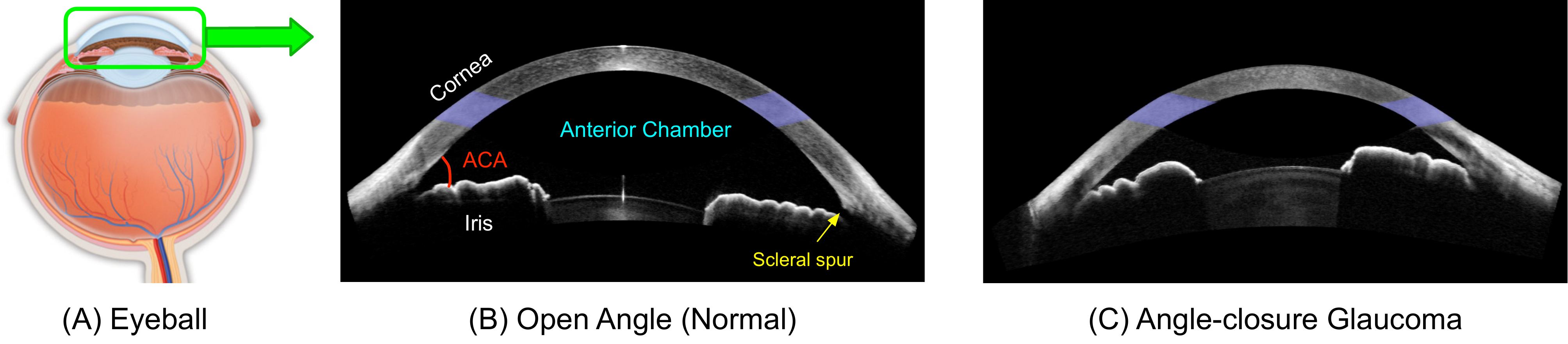}
		\caption{The example of (B) open angle and (C) angle-closure. The narrow anterior chamber angle (ACA) blocks drainage channels of aqueous fluid.}
		\label{img-cover}
	\end{center}
\end{figure}

Glaucoma is the foremost cause of irreversible blindness. Since vision loss from glaucoma cannot be reversed, improved screening and detection methods for glaucoma are essential to preserve vision and life quality. A common type of glaucoma is angle-closure, where the anterior chamber angle (ACA) is narrow as shown in Fig.~\ref{img-cover}, leading to blockage of drainage channels that results in pressure on the optic nerve. Anterior Segment Optical Coherence Tomography (AS-OCT) has been shown to provide an objective method for the evaluation and assessment of ACA structure~\cite{Leung2011}, and thus has been widely used by recent computational techniques for early screening of angle-closure glaucoma.

Recently, several automatic angle-closure glaucoma assessment methods have been studied. Tian \textit{et~al.} provided a Schwalbe’s line detection method for High-Definition OCT (HD-OCT) to compute ACA measurements~\cite{Tian2011}. Xu \textit{et~al.} localized the ACA region, and then extracted visual features directly to classify the glaucoma subtype~\cite{Xu2012,Xu2013}. Fu \textit{et~al.}  proposed a data-driven approach to integrate AS-OCT segmentation, measurement, and screening~\cite{Fu2016EMBC,Fu2017TMI}. These methods segment the AS-OCT image and then extract the representation based on clinical parameters or visual features which are subsequently used for angle-closure classification. These methods, however, lack sufficiently discriminative representations and are easily affected by noise and low quality of the AS-OCT image.
Recently, deep learning has been shown to yield highly discriminative representation that surpass the performance of handcrafted features in many computer vision tasks~\cite{LeCun2015,Gulshan2016}. For example, deep learning has high sensitivity and specificity for detecting referable diabetic retinopathy in retinal fundus photographs~\cite{Gulshan2016}. Deep learning has also been shown to be effective in other retinal fundus applications~\cite{Fu2016,Fu2018DeNet,Fu2018TMI}. These successes have motivated our examination of deep learning for glaucoma assessment in AS-OCT imagery.

A common limitation of deep learning based approaches is the need to downsample the input image to a low resolution (i.e., 224$\times$224) in order for the network size to be computationally manageable~\cite{Gulshan2016}. However, this downsampling leads to a loss of image details that are important for discrimination of subtle pathological changes. To address this issue, we propose to supplement the downsampled global image with a local window chosen at a specific ocular region known to be informative for clinical diagnosis of angle-closure glaucoma~\cite{Nongpiur2013}. The benefit of including this local window is that it is small enough to be processed by a deep network while maintaining the original image details.

Based on this, we propose a Multi-Context Deep Network (MCDN) architecture, which includes two parallel streams that jointly learn predictive representations from the different regions/scales useful for angle-closure estimation. Moreover, an intensity-based data augmentation is utilized to artificially enlarge the AS-OCT training data in order to gain robustness to different AS-OCT imaging devices. The main contributions of this work are as follows: 
(1) We introduce deep learning to the problem of angle-closure glaucoma screening in AS-OCT imagery, for the purpose of gaining more discriminative representations on different regions for glaucoma screening prediction.
(2) A MCDN architecture is developed based on clinical priors about informative image areas. MCDN learns predictive representations for these image regions through processing by separate, parallel network streams.
(3) An intensity-based augmentation of training data is presented to deal with intensity variations among AS-OCT images.
(4) We also demonstrate that incorporating estimated clinical parameters can be beneficial for final glaucoma screening.
(5) A large-scale AS-OCT dataset containing 8270 ACA images is collected for evaluation. Experiments show that our method outperforms the existing screening methods.

\begin{figure*}[!t]
	\begin{center}
		\includegraphics[width=1\linewidth]{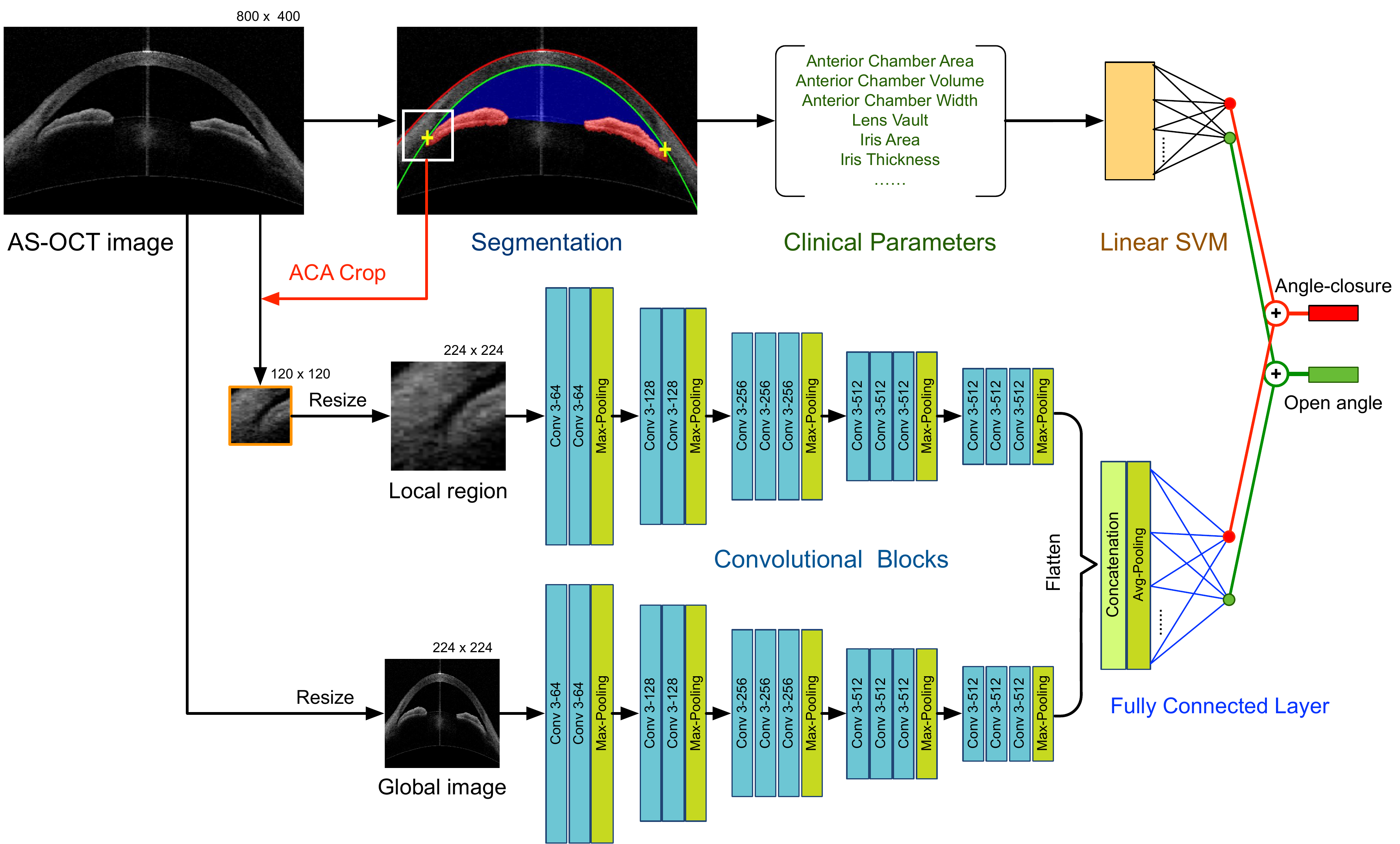}
		\caption{The framework of our angle-closure glaucoma screening system, which contains AS-OCT structure segmentation and MCDN. The CNN layer parameters are denoted as ``Conv (receptive field size)-(number of channels)".}
		\label{img-frame}
	\end{center}
\end{figure*}

\section{Proposed Method}

Our angle-closure glaucoma screening system include two stages, AS-OCT structure segmentation and MCDN, as shown in Fig.~\ref{img-frame}. The AS-OCT structure segmentation is utilized to localize the ACA region and predict the screening result based on clinical parameters, while the MCDN is used to gain the discriminative representation and output the final angle-closure glaucoma screening. Finally, the probability results of clinical parameter and deep learning  are averaged to produce the angle-closure glaucoma screening.

\subsection{AS-OCT Segmentation and Clinical Parameter}

In our system, we implement the data-driven AS-OCT segmentation method in~\cite{Fu2017TMI}, which utilizes marker transfer from labeled exemplars to generate initial markers, and segments the major AS-OCT structure to compute the clinical parameters. A $120 \times 120$ patch centered on the detected ACA is cropped and used as the input of the following MCDN. Moreover, we also calculate the clinical parameters (e.g., Anterior Chamber Width, Lens-Vault, Chamber Height, Iris Curvature, and Anterior Chamber Area) and employ a linear Support Vector Machine (SVM) to predict an angle-closure probability. More details of the clinical parameters can be found in~\cite{Fu2017TMI}.

\subsection{Multi-Context Deep Network Architecture}

Our MCDN architecture includes two parallel streams to obtain the representations for different clinical regions in an AS-OCT image, namely the global fundus image and local disc region. As shown as in Fig.~\ref{img-frame}, each stream consists of a sequence of convolutional units that each contain layers for convolution, batch normalization~\cite{IoffeBN}, and Rectified Linear Units (ReLU) activation~\cite{Krizhevsky2012}. The CNN of each stream learns and extracts local feature representations based on patches randomly sampled from its input. The batch normalization aids training the network~\cite{IoffeBN}, and the ReLU activation function is defined as $f(x) = \max(0, x)$, where $x$ is the input to the neuron.

The first stream processes the cropped patch centered on the ACA region from the AS-OCT segmentation. This area is the most important for clinical diagnosis, as several major clinical measurements are taken from it~\cite{Nongpiur2013} (e.g., iris curvature, angle opening distance, and trabecular-iris space area).
The second stream of our MCDN architecture learns a global feature map for the complete cornea structure. In the clinical domain, the cornea structure offers various cues associated with risk factors for angle-closure glaucoma~\cite{Wu2011}, such as lens vault and anterior chamber width, area and volume.  The input images of the two streams are all resized to $224 \times 224$ to enable use of pre-trained parameters from other deep models~\cite{Krizhevsky2012,Simonyan14c} as the initialization for our network.
The output maps of these two streams are concatenated into a single feature vector that is fed into a fully connected layer to obtain the final screening result. Here, the angle-closure glaucoma screening is formulated as a binary classification task, and softmax regression is employed as a generalization of the logistic regression classifier. The classifier is trained by minimizing the following binary cross-entropy loss function:
\begin{equation}
Loss = - \dfrac{1}{N} \sum_{i=1}^{N} \left\{y_i \log (g(\textbf{w} \cdot \textbf{x}_i)) + (1-y_i) \log (1- g(\textbf{w} \cdot \textbf{x}_i))\right\},
\end{equation}
where $g(\textbf{w} \cdot \textbf{x}_i)$ is the logistic function with weight vector $\textbf{w}$, $(\textbf{x}_i, y_i)$ is the training set containing $N$ AS-OCT images, $\textbf{x}_i$ is the output representation of the $i$-th image and $y_i \in \{0,1\}$ is the angle-closure label.

\subsubsection{Discussion:} A related deep learning model is the Region-based Convolutional Neural Network~\cite{RCNN_2014}, which classifies object proposals using deep convolutional networks. The object proposal regions are obtained using general object detectors, in contrast to our work where we take advantage of clinical guidance to specifically extract the ACA region for glaucoma diagnosis. Our proposed MCDN architecture is also related to multi-scale or multi-column deep networks~\cite{MCNN2012}, which combine features from different scales or receptive field sizes, and generate the final feature map to build a representation. Our network also learns features at different scales, but constructs a more concise representation that focuses on image areas and corresponding scales that are clinically relevant to our problem. Avoiding extraneous information in this way facilitates network training, especially in cases of limited training data. In the following experiment, we also demonstrate that our MCDN architecture outperforms the multi-column network (i.e.,~\cite{MCNN2012}) in the AS-OCT screening task.

Moreover, our system utilizes an individual linear SVM for clinical parameter based estimation instead of concatenating them with deep features into the fully connected layer. This is due to two considerations: 1) the dimensionality of clinical parameters (24 D as in~\cite{Fu2017TMI}) is much lower than that of deep features from the two streams ($2 \times 4096$ D), which would reduce the impact of the clinical parameters; 2) estimation from clinical parameters with linear SVM has demonstrated satisfactory performance in the previous works~\cite{Nongpiur2013,Fu2017TMI}.

\subsection{Data Augmentation for AS-OCT}

AS-OCT images are captured along the perpendicular direction of the eye, and the structure of the anterior chamber appears at a relatively consistent position among AS-OCT images in practice. Traditional image augmentation, e.g. by rotation and scaling, therefore does not aid in AS-OCT screening. On the other hand, image intensities typically vary among different AS-OCT imaging devices, which may affect screening accuracy. We thus employ an intensity-based augmentation to enlarge the data with varied intensities, by rescaling image intensities by a factor $k_I$ ($k_I \in \{ 0.5, 1, 1.5 \}$ in this paper). To increase the robustness of ACA region localization, we additionally perform data augmentation by shifting the ACA position to extract multiple patches as the input to the ACA stream in our MCDN architecture.

\section{Experiments}

For experimentation, we collected a total of 4135 Visante AS-OCT images (Model 1000, Carl-Zeiss Meditec) from 2113 subjects to construct a clinical AS-OCT dataset. Since each AS-OCT image contains two ACA regions, each image is split into two ACA images (8270 ACA images in total), with right-side ACA images flipped horizontally to match the orientations of left-side images. For each ACA image, the ground truth label of open-angle or angle-closure is determined from the majority diagnosis of three ophthalmologists. The data contains 7375 open-angle and 895 angle-closure cases. The dataset is divided equally and randomly into training and testing sets based on subject, such that the two ACAs of one patient will not be separated between the training and test sets.
We employ several evaluation criteria to measure performance:  Sensitivity (Sen), Specificity (Spe), Balanced Accuracy (B-Acc), which are defined as
\begin{eqnarray}
\text{Sen} \text{=}\dfrac{TP}{TP\text{+}FN}, \; \text{Spe} \text{=} \dfrac{TN}{TN\text{+}FP}, \; \text{B-Acc} = \dfrac{1}{2} (\text{Sen} \text{+} \text{Spe}), \nonumber
\end{eqnarray}
where $TP$ and $TN$ denote the number of true positives and true negatives, respectively, and $FP$ and $FN$ denote the number of false positives and false negatives, respectively.  Moreover, we additionally report the ROC curves and area under ROC curve (AUC).

\begin{figure*}[!t]
	\centering
	\includegraphics[width=.48\linewidth]{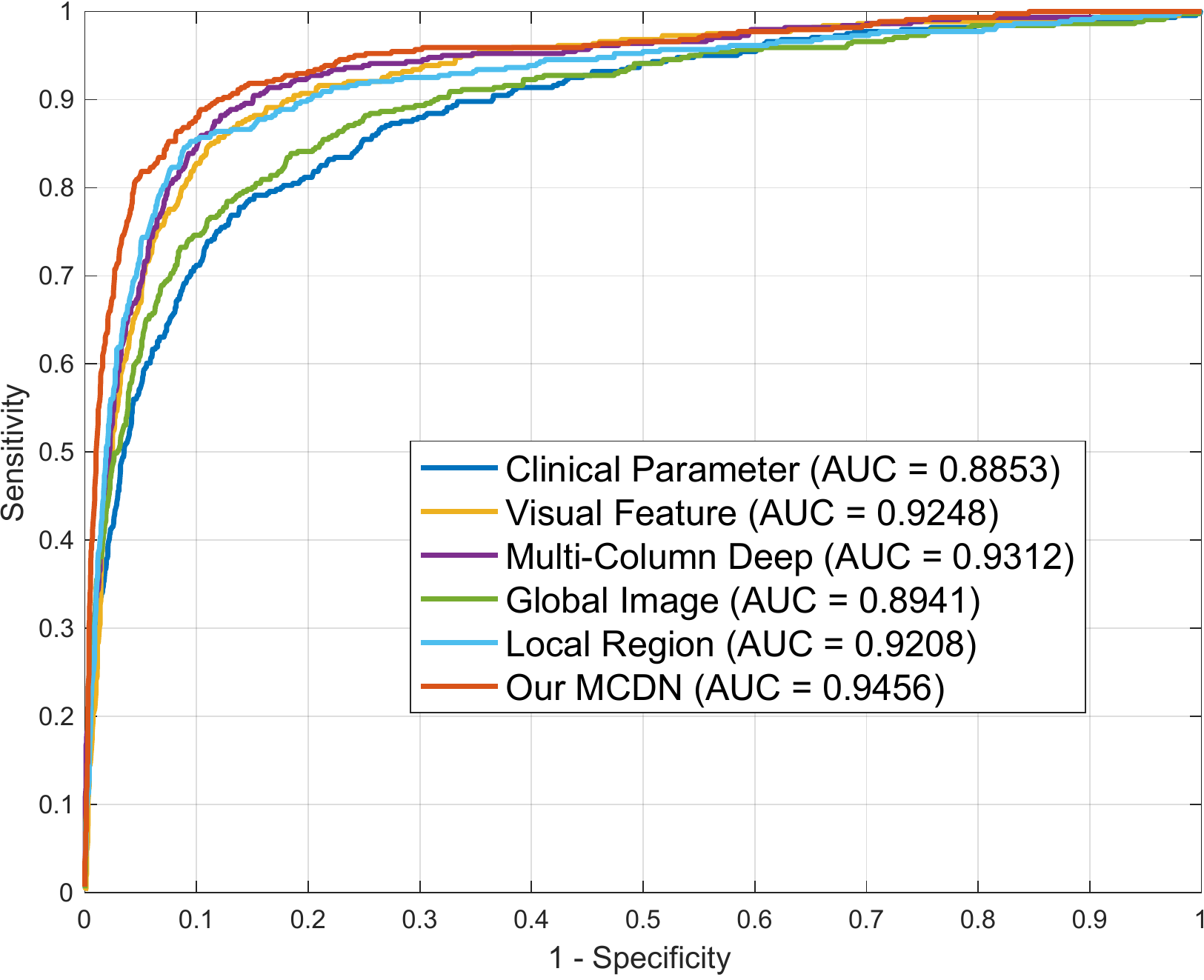} \;
	\includegraphics[width=.48\linewidth]{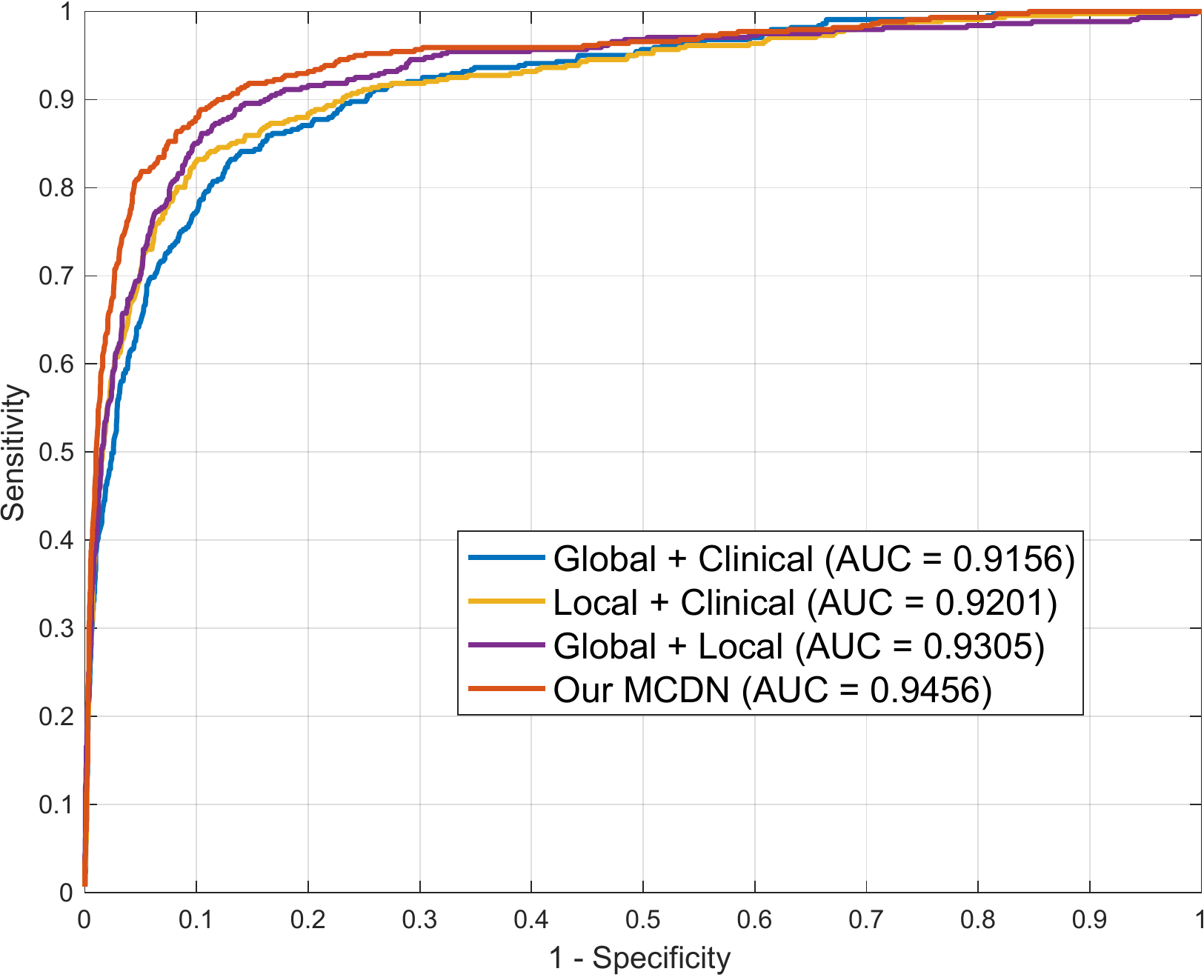}
	\caption {ROC curves with AUC scores on the Visante AS-OCT dataset.}
	\label{img-exp_curve}
\end{figure*}

We compare our algorithm with several AS-OCT screening and deep learning methods: (1) The clinical-based screening method in~\cite{Fu2018TMI}, which segments the AS-OCT image and calculates several clinical parameters. A linear SVM is added to determine a screening result from the parameters. (2) The visual feature-based screening method in~\cite{Xu2013}, which localizes the ACA based on geometric structure and extracts histograms of oriented gradients (HOG) features to classify the glaucoma subtype. The HOG features are computed on a $150 \times 150$ region centered on the ACA, and the classification result is obtained using a linear SVM.
(3) The multi-column deep model in~\cite{MCNN2012}, which is a state-of-the-art deep learning model with different receptive field sizes for each stream. For our algorithm, we show not only the final screening result (Our MCDN), but also provide results obtained without the clinical parameters (Our MCDN w/o CP), and the individual stream result for the global fundus image (Global Image) and for the local disc region (Local Region). The results are reported in Table~\ref{Tab_result} and Fig.~\ref{img-exp_curve}.

\begin{table*}[!t]
    \centering
    \caption{Performance of different methods on the Visante AS-OCT dataset.}
    \begin{tabular}{|C{90pt}|C{57pt}|C{57pt}|C{57pt}|C{57pt}|}
    	\hline
    	                   &       AUC       &   B-Accuracy    &   Sensitivity   &   Specificity   \\ \hline
    	Clinical Parameter &     0.8853      &     0.8198      &     0.7914      &     0.8483      \\
    	  Visual Feature   &     0.9248      &     0.8688      &     0.8503      &     0.8872      \\
    	   Multi-Column    &     0.9312      &     0.8802      &     0.8821      &     0.8783      \\ 
    	   Global Image    &     0.8941      &     0.8286      &     0.7846      &     0.8726      \\
    	   Local Region    &     0.9208      &     0.8790      &     0.8526      & \textbf{0.9055} \\
    	Global + Clinical  &     0.9156      &     0.8508      &     0.8322      &     0.8694      \\
    	 Local + Clinical  &     0.9201      &     0.8657      &     0.8322      &     0.8992      \\
    	  Global + Local   &     0.9305      &     0.8786      &     0.8617      &     0.8955      \\
    	     Our MCDN      & \textbf{0.9456} & \textbf{0.8926} & \textbf{0.8889} &     0.8963      \\ \hline
    \end{tabular}%
    \label{Tab_result}%
\end{table*}%

Clinical parameters are defined by anatomical structure, and most of them have a specific physical significance that clinicians take into consideration in making a diagnosis. By contrast, visual features can represent a wider set of image properties, beyond what clinicians recognize as relevant. Thus visual features perform better than clinical parameters as expected, and achieve $0.9248$ AUC score.   The deep learning based Global Image applied to full AS-OCT images does not work well with only $0.8941$ AUC. A possible reason for this is that although learned discriminative features are more powerful than handcrafted visual features, they are learned in this case over the entire AS-OCT image. By contrast, the Local Region results in performance similar to handcrafted visual features. For the multi-column deep model~\cite{MCNN2012}, the shallow layers can be easier to optimize, giving better results than  visual features. Our MCDN outperforms the other baselines, demonstrating that the proposed fusion of significant clinical regions is effective for angle-closure glaucoma screening. Moreover, we also observe that the combination with clinical parameters leads to an obvious improvement, as the AUC score increases from $0.9305$ to $0.9456$.

\subsubsection{Running Time:} We implement our MCDN system using the publicly available TensorFlow Library. Each stream is fine-tuned from an initialization with the pre-trained VGG-16 deep model in~\cite{Simonyan14c}. The entire fine-tuning phase takes about 5 hours on a single NVIDIA K40 GPU (200 iterations). In testing, it takes 500 ms to output the final screening result for a single AS-OCT image.

\section{Conclusion}

In this paper, we propose an automatic angle-closure glaucoma screening method for AS-OCT imagery via deep learning. A multi-context deep network architecture is proposed to learn discriminative representations on particular regions of different scales. Experiments on a clinical AS-OCT dataset show that our method outperforms the existing screening methods and other state-of-the-art deep architectures. Our MCDN architecture arises from the use of clinical prior knowledge in designing the deep network, and the intensity-based augmentation can also be used in other OCT-based applications.

\bibliographystyle{splncs03}
\bibliography{Deep_ASOCT}

\end{document}